\documentclass{article}
\PassOptionsToPackage{numbers}{natbib}

\usepackage[main,final]{neurips_2026}
\usepackage{graphicx}
\usepackage{amsmath}
\usepackage{enumitem} % Added to support itemize options like [leftmargin=1em]

\usepackage[utf8]{inputenc} % allow utf-8 input
\usepackage[T1]{fontenc}    % use 8-bit T1 fonts
\usepackage{hyperref}       % hyperlinks
\usepackage{url}            % simple URL typesetting
\usepackage{booktabs}       % professional-quality tables
\usepackage{amsfonts}       % blackboard math symbols
\usepackage{nicefrac}       % compact symbols for 1/2, etc.
\usepackage{microtype}      % microtypography
\usepackage{wrapfig}
\usepackage{multirow}
\usepackage[table]{xcolor}  % colors and table row shading
\title{What Remains in Sight? Autoregressive Video Decoding as Representation-Guided Context Rewriting}

\author{%
  Lixuan He \\
  \href{mailto:lixuanhe@stanford.edu}{\texttt{lixuanhe@stanford.edu}} \\
  Stanford University
  \And
  Jihyeon Je \\
  \href{mailto:jihyeonj@stanford.edu}{\texttt{jihyeonj@stanford.edu}} \\
  Stanford University
  \AND
  Yang You\textsuperscript{\ensuremath{\dagger}} \\
  \href{mailto:yangyou@stanford.edu}{\texttt{yangyou@stanford.edu}} \\
  Stanford University
  \And
  Leonidas Guibas\textsuperscript{\ensuremath{\dagger}} \\
  \href{mailto:guibas@stanford.edu}{\texttt{guibas@stanford.edu}} \\
  Stanford University
}

\begin{document}

\maketitle
{\renewcommand{\thefootnote}{\ensuremath{\dagger}}\footnotetext{Corresponding authors.}}

\begin{abstract}
Long-horizon video generation is increasingly moving toward autoregressive rollout, where each newly generated segment becomes part of the evidence used for future prediction. Recent causal decoders, long-context training, and KV-cache mechanisms have extended feasible duration, but most inference pipelines still decide the visible past implicitly through recent windows, compressed caches, or external memory. This raises a direct question: \textbf{can the choice of which past frames or segments to show to the decoder be treated as an explicit test-time decision?} We answer this question with \textsc{ReCR}, a training-free method that formulates autoregressive video decoding as representation-guided context rewriting. At each step, \textsc{ReCR} builds a selected visible context from the generated history by using internal decoder representations to favor context units that match the global rollout state, avoid redundancy with the recent continuation, preserve temporal boundary coverage, and widen the range of spatial layouts in view. The selected units are then mapped onto a compact causal axis before decoding the next block. Across long-horizon text-to-video benchmarks, \textsc{ReCR} consistently improves multiple autoregressive backbones under matched visible-token budgets. Empirical analyses such as representation-space diagnostics and transition-count prompt switching further show that improving what remains visible is an effective path toward more stable long-video decoding. Our code is available at \url{https://github.com/lixuan27/ReCR}.
\end{abstract}

\section{Introduction}
\label{sec:introduction}

Long-horizon video generation is moving from short, self-contained clips toward continuous and controllable synthesis over much longer durations. Autoregressive video models are especially suited to this setting because they generate new content by repeatedly conditioning on previously generated content. Recent progress has improved this direction through self-rollout training, causal distillation, next-frame prediction, long-context supervision, semi-autoregressive block generation, and horizon-extension strategies \citep{huang2025selfforcing,cui2025selfforcingpp,liu2025infinitystar,cheng2025adaptivebov,yang2025onestepcausal,yesiltepe2025infinityrope,zhang2025blockvid,chen2025recurrentar,yu2025videossm,yi2025deepforcing,ji2026videoar,zhu2026causalforcing,chen2026contextforcing,li2026flex}. These methods show that causal video generation can scale beyond the short-clip setting, while also making a key failure mode more visible: as generation proceeds, the model repeatedly reads its own past, so any imbalance in the visible context can propagate into identity drift, scene inconsistency, motion decay, or delayed response to new directives.

A large part of recent work has addressed this difficulty by improving how long histories are stored, compressed, or accelerated. Memory-based methods introduce explicit history encoders, state-space memory, entity-level memory banks, or recurrent attention to preserve long-range consistency \citep{yu2025videossm,zhu2025mag,zhou2026videomemory,ghafoorian2026rehyat,zhang2025framepreserve}. System-oriented methods reduce the cost of long histories through KV-cache compaction, quantization, token pruning, cache reuse, sparse attention, and block-level attention caching \citep{li2026packcache,chen2026pafukv,xi2026qvg,guo2026dummyforcing,samuel2026fastarcache,lv2026lightforcing,shaulov2026tokentrim,chen2026flashblock,ma2026flowcache,agarwal2026monarchrt}. These advances are essential for efficient long video synthesis, but they mainly ask how to make a long context cheaper or more stable. They leave a more direct inference-time question underexplored: under a fixed visible context budget, how should the decoder choose which parts of the generated past to read for the next prediction?

\begin{figure}
    \centering
    \includegraphics[width=\linewidth]{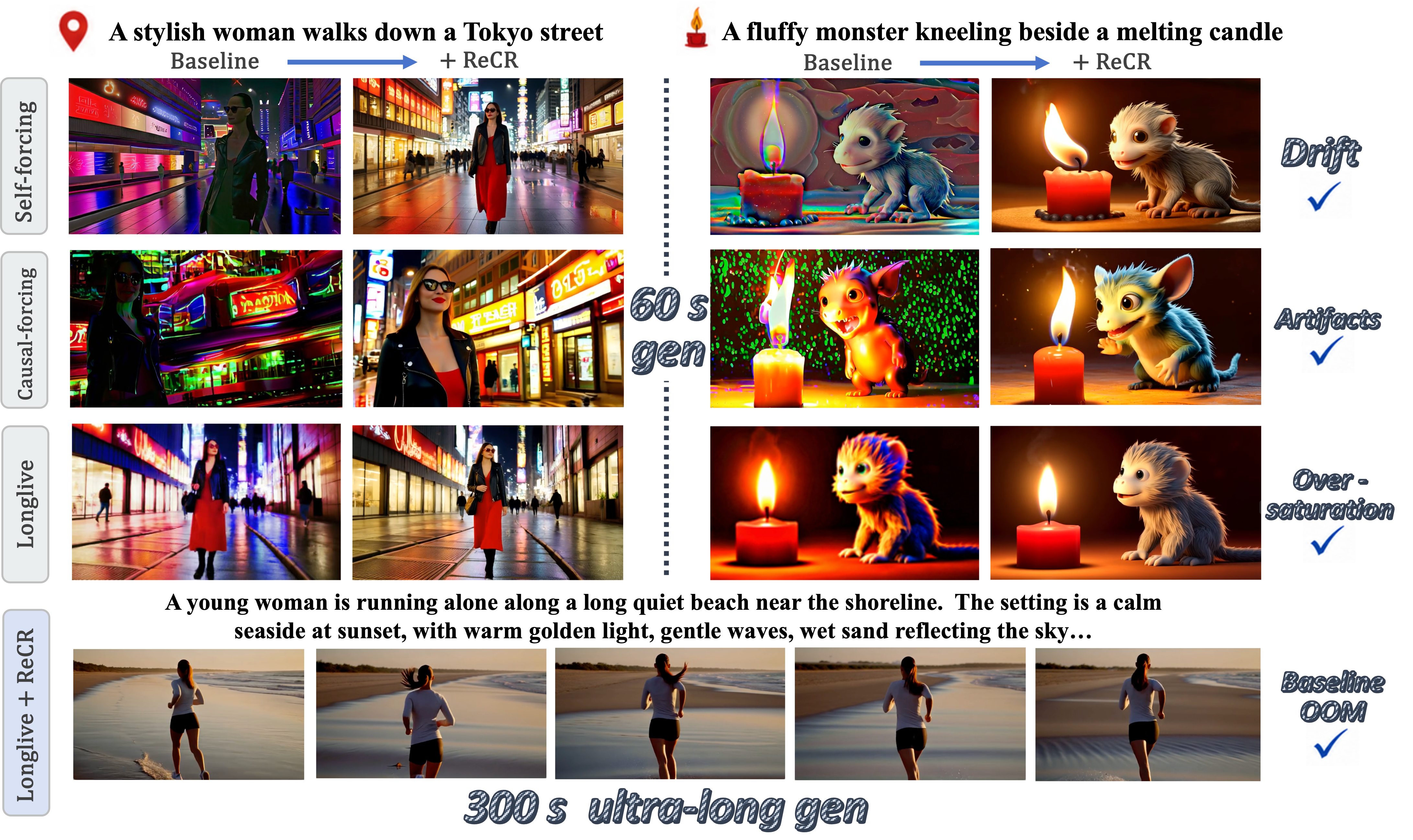}
    \caption{\textbf{Qualitative overview of ReCR across backbones.}
    ReCR is applied as a training-free context rewriting module on top of different autoregressive video backbones. On \(60\)-second rollouts, ReCR reduces common long-horizon failures such as scene drift, visual artifacts, and oversaturation. On \(300\)-second ultra-long rollouts, ReCR further extends backbones with stable visual continuity.}
    \label{fig:result}
\end{figure}

This question is important because the useful past in a long video is unlikely to be uniformly determined by recency. A recent window is often necessary for local temporal continuity, yet it may overemphasize the immediate continuation and underuse earlier evidence that stabilizes identity, layout, or object state. Conversely, retaining the full context can be expensive and can introduce redundant or stale evidence. Recent studies on history encoding, unstable-token pruning, reward feedback, adaptive sampling, prompt switching, and controllable autoregressive generation all suggest that long-horizon failures often depend on which trajectory evidence is reused during inference \citep{yesiltepe2025infinityrope,zhang2026rewardforcing,han2026entropyguard,shaulov2026tokentrim,xiao2025knotforcing,fosdick2026vace,wang2026tagrpo,po2025bagger,qu2025ttom,yuan2026physicsalign,wang2026mechanisticview}. We therefore treat context choice itself as the object of study. The central problem is to directly construct a useful visible context from the generated history at every rollout step, rather than relying only on chronological recency, full-history retention, or cache-level compression.

We propose \textsc{ReCR}, a training-free decoding strategy for representation-guided context rewriting. At each rollout step, ReCR uses internal decoder representations to score candidate context units, selects a sparse visible context under a fixed budget, refines it with at most one spatial-range exchange, and presents the selected units on a compact causal axis before decoding the next block. This design is complementary to long-context training and KV-cache methods: those approaches mainly improve the model, the memory system, or the efficiency of retaining history, while ReCR changes the composition of the context that the decoder actually reads. Across standard long-video benchmarks, drift diagnostics, representation-space analysis, and transition-count prompt switching, ReCR improves long-horizon stability and motion evolution under matched visible-token budgets.
Our contributions are threefold:

\begin{itemize}[leftmargin=1em]
    \item We formulate long-horizon autoregressive video decoding as a context selection problem, where the key inference-time decision is to directly choose a sparse visible context from the generated history under a fixed budget.
    \item We propose \textsc{ReCR}, a training-free context rewriting method that uses internal representations to select useful context units, balance global compatibility with recent-context redundancy, and present the selected units on a compact causal axis.
    \item We provide a mechanistic evaluation of visible context construction, including dynamic context composition, representation-space diagnostics, and transition-count prompt switching, showing that ReCR improves evidence allocation beyond contiguous recent context.
\end{itemize}

\section{Related Work}
\label{sec:related_work}

\subsection{Video diffusion models and foundation-scale generation}

Modern video generation was first driven by full-sequence diffusion and transformer-based formulations, which use bidirectional spatiotemporal denoising to produce visually strong short clips \citep{peebles2022dit,ho2022videodiffusion,ho2022imagenvideo,guo2023animatediff,blattmann2023stablevideo,ma2024latte,qiu2023freenoise,zhai2024mcm,kondratyuk2023videopoet,yang2024cogvideox}. This line then scaled toward foundation-scale text-to-video systems, unified generation and editing frameworks, stronger video VAEs, larger curated datasets, and broader evaluation \citep{tencent2024hunyuanvideo,opensoraplan2024,peng2025opensora2,polyak2024moviegen,wanteam2025wan,ju2025fulldit,jiang2025vace,ma2025stepvideo,chen2025skyreelsv2,zheng2025vbench2}. These advances improve visual fidelity, prompt following, motion realism, and controllability, but long-duration generation remains challenging because full-sequence denoising scales poorly and typically assumes a fixed offline clip. Building on this progress, ReCR asks which generated context should remain visible for the next prediction.

\subsection{Causal and autoregressive long-video decoding}

A second line of work reframes video generation as causal rollout, enabling streaming synthesis, interactive prompting, and minute-scale extension through diffusion distillation, next-frame or next-chunk prediction, self-rollout training, and long-context supervision \citep{chen2024diffusionforcing,yin2024causvid,henschel2024streamingt2v,gu2025far,huang2025selfforcing,yang2025longlive,liu2025rollingforcing,wu2025packforce,huang2025memoryforcing,cui2025selfforcingpp,jia2025moga,shin2025motionstream,chen2025recurrentar,liu2025infinitystar,cheng2025adaptivebov,yang2025onestepcausal,zhang2025blockvid,ji2026videoar,zhu2026causalforcing,chen2026contextforcing,zou2026hiar,liu2026diagdistill,he2026arcopo,luo2026shotstream}. These methods make long video generation more compatible with KV caching and online control, and several recent works explicitly target horizon extension, prompt switching, multi-shot consistency, or stronger training-inference alignment \citep{yesiltepe2025infinityrope,yi2025deepforcing,li2026flex,xiao2025knotforcing,luo2025filmweaver,zhu2025astra}. ReCR complements them by keeping the decoder fixed and directly selecting the visible context under the same budget.

\subsection{Memory, cache efficiency, and test-time inference for long videos}

Long-horizon generation has also been studied through memory management, cache compression, sparse attention, and test-time optimization. Memory-based methods introduce state-space memory, recurrent memory, and spatial retrieval to improve long-range consistency \citep{yu2025videossm,ji2025memflow,zhu2025mag,zhou2026videomemory,ghafoorian2026rehyat,zhang2025egolcd,hong2025relic,zhang2025storymem,wu2026infiniteworld,chen2026hybridmemory}. System-oriented methods reduce the cost of retaining history with KV-cache compaction, quantization, token pruning, attention caching, sparse attention, and redundancy-eliminated streaming \citep{zhang2025framepreserve,li2026packcache,chen2026pafukv,xi2026qvg,guo2026dummyforcing,samuel2026fastarcache,lv2026lightforcing,shaulov2026tokentrim,chen2026flashblock,ma2026flowcache,agarwal2026monarchrt,qiu2025histream,li2025radialattention,luo2026svoo}. In parallel, test-time training, test-time search, reward feedback, trajectory correction, and adaptive sampling show that inference computation can be used to improve generated trajectories without changing the base model \citep{dalal2025oneminute,liu2025videot1,he2025evosearch,wu2025imagerysearch,zhang2026rewardforcing,han2026entropyguard,wang2026tagrpo,qu2025ttom,huang2025sketchverify,yuan2026physicsalign,po2025bagger,wang2026mechanisticview}. ReCR does not compress cache, train longer contexts, or search over output candidates, but directly constructs the visible context.

\begin{figure}
    \centering
    \includegraphics[width=\linewidth]{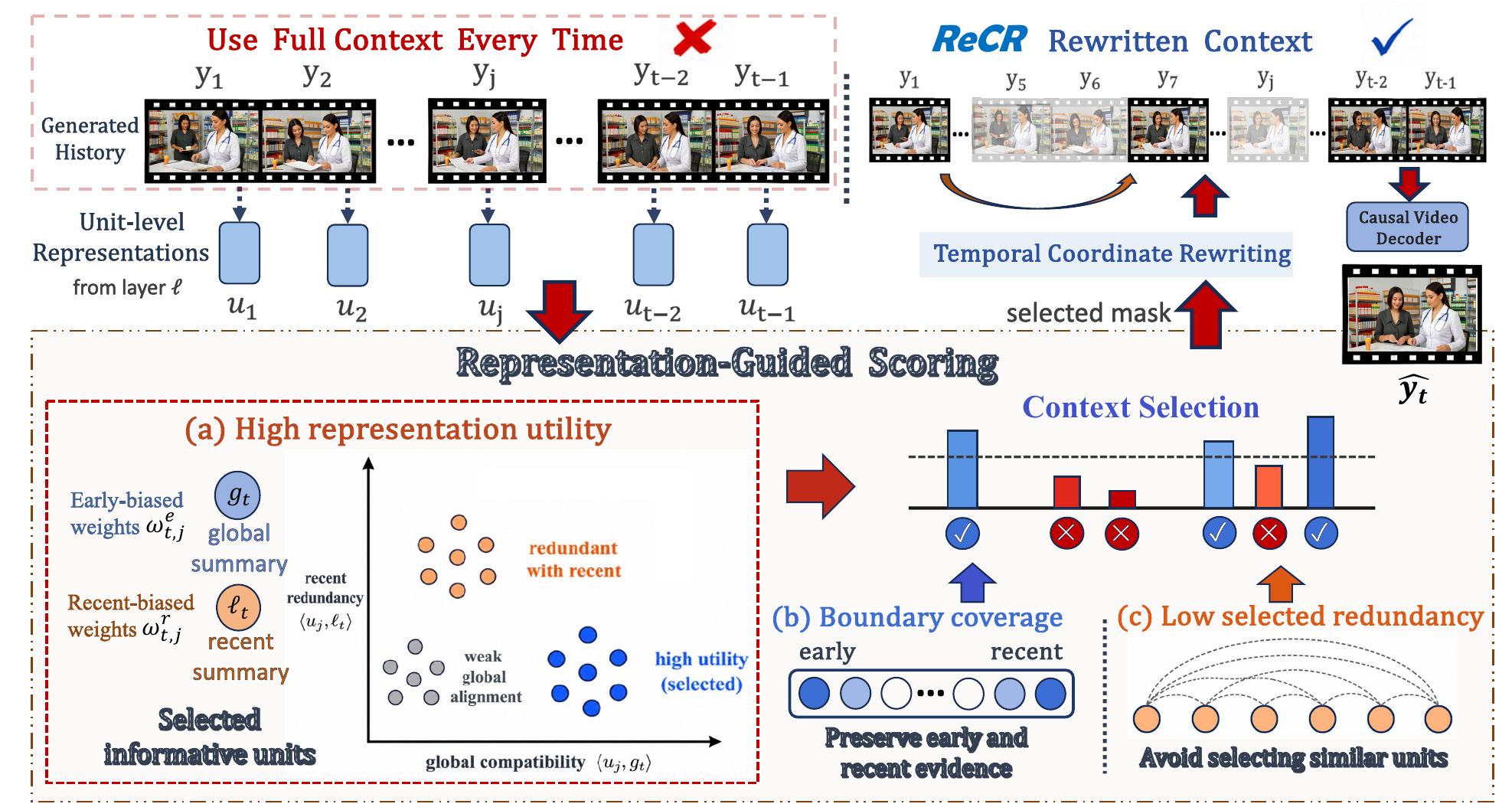}
    \caption{\textbf{Overview of the ReCR framework.}
    Instead of exposing the full generated history or a fixed recent window, ReCR scores context units with unit-level representations and selects a compact visible context. The selection objective favors high representation utility, preserves boundary coverage over early and recent context, and discourages redundant selected units. At most one spatial-range exchange, not drawn here, then refines the selection (Sec.~\ref{subsec:method_selection}). The selected context is then temporally reindexed onto a compact causal axis before being passed to the causal video decoder.}
    \label{fig:teaser}
\end{figure}

\section{Method}
\label{sec:method}

We study autoregressive video decoding from the viewpoint of representation-guided context rewriting. At rollout step \(t\), the model has produced a generated history \(H_t\). Standard decoding exposes mainly a contiguous recent context \(W_t\). This choice leaves open a question: which context units from \(H_t\) should remain visible when the decoder predicts the next block?
ReCR answers this question by constructing a selected visible context \(C_t\) at each step. The selected units are scored with internal unit representations, chosen under a fixed context budget, and placed on a compact causal axis through a temporal reindexing map \(\pi_t\). The method changes the context presented to the decoder while keeping the backbone parameters fixed.

\subsection{Autoregressive Decoding as Context Rewriting}
\label{subsec:method_formulation}

\paragraph{Setup.}
Let a video be decomposed into latent frames \(y_1,y_2,\ldots,y_T\), which we call context units. The decoder generates them in consecutive blocks of \(L\) units, with \(L=3\) for all backbones in this paper, so rollout steps are the block starts \(t\in\mathcal{T}=\{1,L+1,2L+1,\ldots\}\). Let \(c\) be the text condition and let \(\theta\) be the parameters of a causal video decoder. Before predicting the block \(y_{t:t+L-1}\), the generated history is \(H_t=\{y_1,y_2,\ldots,y_{t-1}\}\). A standard autoregressive decoder conditions on a contiguous recent context \(W_t=\{y_j \mid a_t \le j < t\}\), where \(a_t\) is a start index. This form assumes that recency is the main proxy for usefulness. In long-horizon rollout, this assumption can be incomplete: the useful evidence for the next prediction may include early context units, recent context units, and intermediate context units. Their roles are therefore treated as an empirical and algorithmic question rather than as a fixed prior.

ReCR replaces the fixed context rule with a step-dependent selected visible context \(C_t \subseteq H_t\). The decoder predicts with
\[
p_{\theta}(y_{1:T} \mid c)
=
\prod_{t\in\mathcal{T}}
p_{\theta}(y_{t:t+L-1} \mid c, C_t, \pi_t),
\]
where \(\pi_t\) is a temporal reindexing map that presents the selected context units on a compact causal axis.

\paragraph{Objective.}
Let \(B\) be the context budget, counted in history units and excluding the \(L\) units of the current block, and define the feasible set \(\Omega_t=\{C \subseteq H_t \mid |C|=\min(B,t-1)\}\). The ideal choice is \(C_t^{\star}=\arg\max_{C\in\Omega_t}J_t(C)\), where \(J_t(C)\) measures the usefulness of \(C\) for predicting the current block. Since \(J_t\) is not observed at inference time, ReCR builds a surrogate from the decoder's own representations. The surrogate should capture three desiderata. First, selected units should agree with the global state summarized from \(H_t\). Second, they should add information beyond the immediate recent continuation. Third, the selected set should keep enough temporal boundary coverage so that early and recent evidence are not both lost during sparse selection.

\subsection{Representation-Guided Context Selection}
\label{subsec:method_selection}

\paragraph{Unit representation.}
ReCR first assigns each context unit a reusable representation. Fix a decoder layer \(\ell\). For unit \(y_j\), let \(Z_j^{(\ell)}=\{z_{j,m}^{(\ell)}\in\mathbb{R}^{d}\mid 1\le m\le n_j\}\) be the output tokens of that layer for \(y_j\) in the clean context pass, which re-encodes a finished block at noise level zero to write it to the decoder cache. The unit representation is
\[
u_j^{(\ell)}
=
\operatorname{Norm}
\left(
\frac{1}{n_j}
\sum_{m=1}^{n_j}
z_{j,m}^{(\ell)}
\right),
\qquad
\operatorname{Norm}(v)
=
\frac{v}{\lVert v \rVert_2}.
\]
We write \(u_j\) when the layer is clear. Each \(u_j\) is computed once at commit time and reused at later steps, so ReCR adds no decoder forward pass.

\paragraph{Global and recent summaries.}
To obtain coarse views of \(H_t\), define the normalized temporal coordinate \(\tau_t(j)=(j-1)/(t-2)\) for \(1\le j<t\). ReCR uses an early-biased weight and a recent-biased weight,
\[
\omega_{t,j}^{e}
=
\frac{\exp(-\beta \tau_t(j))}
{\sum_{q<t}\exp(-\beta \tau_t(q))},
\qquad
\omega_{t,j}^{r}
=
\frac{\exp(-\beta (1-\tau_t(j)))}
{\sum_{q<t}\exp(-\beta (1-\tau_t(q)))}.
\]
They induce the global summary \(g_t=\operatorname{Norm}\left(\sum_{j<t}\omega_{t,j}^{e}u_j\right)\) and the recent summary \(\ell_t=\operatorname{Norm}\left(\sum_{j<t}\omega_{t,j}^{r}u_j\right)\).
The summaries do not assume that early or recent context units are always sufficient. They provide two reference directions for scoring each candidate unit in \(H_t\).

\begin{figure}
    \centering
    \includegraphics[width=\linewidth]{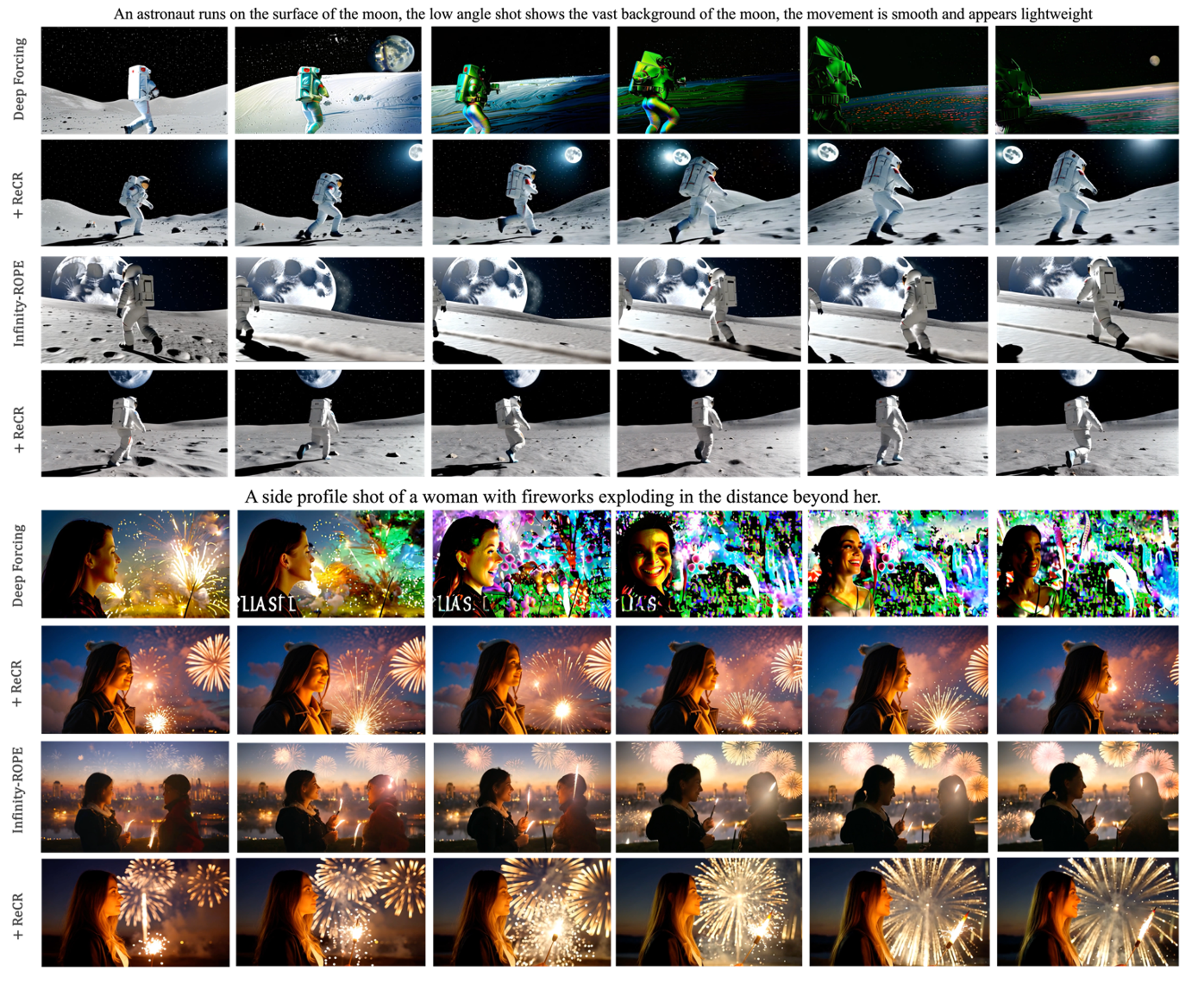}
    \caption{\textbf{ReCR further strengthens training-free decoding methods.}
    Qualitative comparisons on two \(60\)-second prompts. For both methods, ReCR improves temporal consistency and visual fidelity over time: it reduces drift and artifacts in Deep Forcing, and further improves the stability and text-alignment ability of Infinity-RoPE. These results show that ReCR is a complementary context-rewriting module that can be added on top of strong training-free methods.}
    \label{fig:result_training_free}
\end{figure}

\paragraph{Representation utility.}
The representation utility of \(y_j\) is
\[
s_t(j)
=
\kappa \langle u_j, g_t \rangle
-
\lambda
\log
\left(
\varepsilon
+
\exp(\kappa \langle u_j, \ell_t \rangle)
\right),
\]
where \(\varepsilon>0\) is a small numerical constant. The first term measures compatibility with the global summary. The second term penalizes redundancy with the recent summary. A high score therefore indicates that a context unit is aligned with the broader rollout state while adding evidence beyond the recent continuation.

\paragraph{Budgeted context selection.}
Let \(m_{t,j}\in\{0,1\}\) indicate whether \(y_j\) is selected for \(1\le j<t\). ReCR forms an initial selection \(\widetilde{C}_t\) from
\[
m_t^{\star}
=
\arg\max_{m_t}
\left[
\sum_{j<t} m_{t,j}s_t(j)
+
\mu \Phi_t(m_t)
-
\gamma \Psi_t(m_t)
\right]
\quad
\mathrm{s.t.}
\quad
\sum_{j<t}m_{t,j}=\min(B,t-1).
\]
Here
\[
\Phi_t(m_t)
=
\log
\left(
\varepsilon
+
\sum_{j<t}m_{t,j}\omega_{t,j}^{e}
\right)
+
\log
\left(
\varepsilon
+
\sum_{j<t}m_{t,j}\omega_{t,j}^{r}
\right)
\]
is the boundary coverage term, and
\[
\Psi_t(m_t)
=
\sum_{1 \le i < j < t}
m_{t,i}m_{t,j}
\langle u_i,u_j\rangle
\]
is the selected-set redundancy penalty. The initial selection is \(\widetilde{C}_t=\{y_j\in H_t \mid m_{t,j}^{\star}=1\}\), computed with the greedy approximation of App.~\ref{app:recr_greedy}; when \(t-1\le B\) it is the whole history. This objective lets the composition of \(\widetilde{C}_t\) emerge from the representation geometry of the current rollout. Boundary coverage keeps early and recent context units available under the budget. Representation utility and the selected-set redundancy penalty determine which additional context units are useful, including intermediate context units when they provide evidence for event evolution.

\paragraph{Spatial-range exchange.}
Mean pooling in \(u_j\) discards how content is laid out within a frame. ReCR therefore tracks a scalar spatial-layout statistic \(\rho_j\), the size of the first-order spatial moments of the layer-\(\ell\) tokens relative to their mean (App.~\ref{app:spatial_exchange}), and refines \(\widetilde{C}_t\) with at most one exchange per step. When \(t-1>B\) and the latest \(L\) units are all selected, the history extreme of \(\rho_j\) lying farthest outside the range covered by \(\widetilde{C}_t\) replaces the lowest-utility selected unit that is neither one of the latest \(L\) units nor temporally adjacent to another selected unit, provided that the covered range does not shrink. The result is the selected visible context \(C_t\), with the same size as \(\widetilde{C}_t\); if no such pair exists, \(C_t=\widetilde{C}_t\).

\subsection{Temporal Reindexing and Decoding}
\label{subsec:method_reindex}

The selected visible context \(C_t\) is usually sparse in the generated history \(H_t\). Directly feeding the original sparse positions can make the causal decoder interpret large temporal gaps as part of the visible sequence. ReCR therefore applies temporal reindexing.

Let \(j_t^{(1)} < j_t^{(2)} < \cdots < j_t^{(N_t)}\) be the selected indices in chronological order, where \(C_t=\{y_{j_t^{(1)}},y_{j_t^{(2)}},\ldots,y_{j_t^{(N_t)}}\}\) and \(N_t=|C_t|\). The temporal reindexing map is \(\pi_t\big(y_{j_t^{(i)}}\big)=i\) for \(1\le i\le N_t\), and the current block follows at positions \(N_t+1,\ldots,N_t+L\).
The selected units keep their chronological order and are presented on a compact causal axis: ReCR stores keys and values before positional encoding and, at each use, applies RoPE to the selected keys at these compact positions. Let \(\xi_t\) denote the decoder input associated with predicting the current block. The final decoding rule is
\[
\hat{y}_{t:t+L-1}
=
D_{\theta}
\big(
\xi_t, c, C_t, \pi_t
\big).
\]
ReCR therefore consists of four operations: compute unit representations from \(H_t\), select \(\widetilde{C}_t\) with representation utility, boundary coverage, and the selected-set redundancy penalty, refine it into \(C_t\) with the spatial-range exchange, and decode after temporal reindexing. \(C_t\) and \(\pi_t\) are computed once per block and shared by all decoder layers and all generator passes of that block. Additional implementation details and the explicit inference procedure are provided in App.~\ref{app:recr_algorithm}.

\begin{table*}[t!]
\centering
\caption{VBench-Long evaluation on the \(30\)- and \(60\)-second benchmarks, reported as \(30\mathrm{s}/60\mathrm{s}\). Higher is better. Best and second-best values are highlighted in bold and underlined, respectively.}
\label{tab:vbench_30s_60s_merged}
\setlength\tabcolsep{7.2pt}
\renewcommand{\arraystretch}{1.20}
\small
\resizebox{\textwidth}{!}{%
\begin{tabular}{lcccccccc}
\toprule
\multirow{2}{*}{\textbf{Method}} &
\multicolumn{8}{c}{\textbf{Quality block: 30s / 60s}} \\
\cmidrule(lr){2-9}
& \shortstack[c]{\textbf{Total}\\\textbf{Score}}
& \shortstack[c]{\textbf{Dynamic}\\\textbf{Degree}}
& \shortstack[c]{\textbf{Subject}\\\textbf{Consistency}}
& \shortstack[c]{\textbf{Background}\\\textbf{Consistency}}
& \shortstack[c]{\textbf{Temporal}\\\textbf{Flickering}}
& \shortstack[c]{\textbf{Motion}\\\textbf{Smoothness}}
& \shortstack[c]{\textbf{Aesthetic}\\\textbf{Quality}}
& \shortstack[c]{\textbf{Imaging}\\\textbf{Quality}} \\
\midrule
\multicolumn{9}{l}{\textbf{Training-based Baselines}} \\
\midrule
Self-Forcing \citep{huang2025selfforcing}
& 79.09 / 78.19
& 40.12 / 33.45
& 96.75 / 96.67
& 95.63 / 95.82
& 95.93 / 96.02
& 96.31 / 97.17
& 56.01 / 54.21
& 70.49 / 66.44 \\

Self-Forcing++ \citep{cui2025selfforcingpp}
& 79.56 / 79.59
& 39.45 / 38.67
& 96.71 / 96.43
& 95.76 / 95.96
& 95.10 / 96.63
& 97.83 / 97.25
& 56.64 / 56.13
& 70.15 / 69.20 \\

LongLive \citep{yang2025longlive}
& 81.32 / 80.72
& 36.74 / 35.43
& 97.63 / 97.42
& \underline{96.15} / \underline{96.08}
& \textbf{97.77} / 97.21
& 98.25 / 98.17
& \underline{58.82} / 58.21
& 70.57 / 70.02 \\

Causal-Forcing \citep{zhu2026causalforcing}
& 80.77 / 77.41
& \underline{62.64} / 56.82
& 95.78 / 94.86
& 95.30 / 95.27
& 96.10 / 95.20
& 97.64 / 96.44
& 54.43 / 48.22
& 68.35 / 63.26 \\

HiAR \citep{zou2026hiar}
& 79.76 / 79.36
& 32.83 / 30.34
& 97.62 / \underline{97.62}
& 95.12 / 94.64
& 96.48 / 96.48
& 98.11 / 98.35
& 56.25 / 55.41
& 70.27 / 69.63 \\
\midrule
\multicolumn{9}{l}{\textbf{Training-free Methods}} \\
\midrule
Deep Forcing \citep{yi2025deepforcing}
& 81.96 / 81.42
& 55.21 / 55.45
& 97.01 / 96.21
& \underline{96.15} / 96.05
& 96.25 / 95.73
& 98.18 / 98.18
& \textbf{60.15} / \textbf{59.27}
& 69.22 / 68.92 \\

Infinity-RoPE \citep{yesiltepe2025infinityrope}
& 80.93 / 80.98
& 45.21 / 45.04
& 97.19 / 97.43
& 96.04 / 95.94
& 97.14 / \underline{97.45}
& 98.09 / 98.32
& 58.08 / 56.88
& 67.43 / 67.25 \\
\midrule
\multicolumn{9}{l}{\textbf{Backbone + Ours}} \\
\midrule
\rowcolor{gray!12}
Self-Forcing + ReCR
& 82.05 / 81.60
& 55.83 / 56.88
& 96.69 / 96.43
& 95.99 / 95.76
& 96.78 / 96.68
& 98.23 / 98.08
& 58.05 / 56.99
& 70.58 / 69.61 \\

\rowcolor{gray!12}
Self-Forcing++ + ReCR
& 81.74 / 81.15
& 57.37 / 56.03
& 96.56 / 96.38
& 95.83 / 95.50
& 96.85 / 96.44
& 97.41 / 97.07
& 58.55 / 58.48
& 70.25 / 70.14 \\

\rowcolor{gray!12}
LongLive + ReCR
& 81.60 / 81.37
& 37.29 / 38.54
& \textbf{97.70} / 97.60
& \textbf{96.52} / \textbf{96.35}
& \underline{97.76} / \textbf{97.71}
& \textbf{98.90} / \textbf{98.86}
& 57.87 / 57.29
& 70.28 / 69.31 \\

\rowcolor{gray!12}
Causal-Forcing + ReCR
& \textbf{82.50} / \underline{82.08}
& \textbf{77.97} / \textbf{80.47}
& 95.67 / 95.38
& 95.15 / 94.84
& 95.48 / 95.14
& 97.77 / 97.58
& 56.74 / 55.64
& \underline{71.15} / \underline{70.62} \\

\rowcolor{gray!12}
HiAR + ReCR
& 80.77 / 80.71
& 33.75 / 35.55
& \underline{97.68} / \textbf{97.75}
& 96.07 / 96.07
& 97.05 / 97.01
& 98.30 / 98.29
& 57.65 / 56.76
& \textbf{71.44} / \textbf{71.08} \\
\midrule
\multicolumn{9}{l}{\textbf{Backbone Training-free Methods + Ours}} \\
\midrule
\rowcolor{gray!12}
Deep Forcing + ReCR
& \underline{82.36} / \textbf{82.27}
& 58.39 / \underline{58.07}
& 96.72 / 96.56
& 96.01 / 95.88
& 96.73 / 96.84
& 98.24 / 98.30
& 58.44 / \underline{58.64}
& 70.96 / 70.17 \\

\rowcolor{gray!12}
Infinity-RoPE + ReCR
& 82.23 / 82.07
& 55.31 / 57.14
& 97.22 / 97.13
& 96.14 / 95.99
& 97.17 / 97.03
& \underline{98.51} / \underline{98.42}
& 58.53 / 57.89
& 68.67 / 68.35 \\
\bottomrule
\end{tabular}%
}
\end{table*}

\begin{table*}[t!]
\centering
\caption{Drift metrics on MovieGen-128 \(30\)-second videos. Best and second-best values in each column are highlighted in bold and underlined, respectively.}
\label{tab:drift_abs_30s}
\setlength\tabcolsep{4.2pt}
\renewcommand{\arraystretch}{1.10}
\small
\resizebox{\textwidth}{!}{%
\begin{tabular}{lcccccccc}
\toprule
\multirow{2}{*}{\textbf{Method}} &
\multicolumn{2}{c}{\textbf{Semantic stability}} &
\multicolumn{4}{c}{\textbf{Appearance / low-level drift}} &
\multicolumn{2}{c}{\textbf{Temporal drift}} \\
\cmidrule(lr){2-3}\cmidrule(lr){4-7}\cmidrule(lr){8-9}
& \textbf{TA overall} $\uparrow$
& \textbf{TA tail} $\downarrow$
& \textbf{Aes. drift} $\downarrow$
& \textbf{AQ signed} $\to 0$
& \textbf{SAT shift} $\to 0$
& \textbf{SHARP tail} $\to 0$
& \textbf{LPIPS tail} $\to 0$
& \textbf{DINO tail} $\to 0$ \\
\midrule
\multicolumn{9}{l}{\textbf{Training-based Baselines}} \\
\midrule
Self-Forcing baseline
& 24.14 & +2.47 & 5.81 & -3.61 & +5.76 & -335.23 & -4.74 & -0.31 \\

LongLive baseline
& 26.45 & \underline{+0.37} & 3.05 & -1.01 & -2.11 & +42.55 & +1.30 & +0.45 \\

Causal-Forcing baseline
& 23.59 & +4.11 & 9.70 & -8.66 & +7.02 & -316.66 & -3.18 & +1.64 \\

HiAR baseline
& 26.27 & +0.45 & 3.67 & -1.78 & +1.72 & -259.35 & +1.52 & \underline{+0.09} \\
\midrule
\multicolumn{9}{l}{\textbf{Training-free Methods}} \\
\midrule
Deep Forcing
& 25.69 & +0.84 & 3.56 & -1.60 & -1.74 & -44.67 & +2.07 & +0.93 \\

Infinity-RoPE
& 25.92 & +0.64 & 3.30 & -1.48 & +0.64 & -9.12 & \underline{-1.26} & +0.12 \\
\midrule
\multicolumn{9}{l}{\textbf{Backbone + Ours}} \\
\midrule
\rowcolor{gray!12}
Self-Forcing + ReCR
& 26.71 & +1.00 & 3.99 & -1.74 & -1.44 & \textbf{+2.51} & \textbf{+0.82} & +0.72 \\

\rowcolor{gray!12}
LongLive + ReCR
& \textbf{27.27} & +1.05 & 3.27 & -2.20 & -2.21 & +44.06 & +2.47 & +0.37 \\

\rowcolor{gray!12}
Causal-Forcing + ReCR
& 25.83 & +1.50 & 5.13 & -4.24 & +3.41 & -185.74 & +2.89 & +1.39 \\

\rowcolor{gray!12}
HiAR + ReCR
& \underline{27.08} & \textbf{+0.27} & 3.02 & -0.90 & +1.29 & -287.90 & +1.40 & \textbf{-0.08} \\
\midrule
\multicolumn{9}{l}{\textbf{Backbone Training-free Methods + Ours}} \\
\midrule
\rowcolor{gray!12}
Deep Forcing + ReCR
& \underline{27.08} & +0.39 & \textbf{1.76} & \textbf{-0.09} & \underline{-0.49} & \underline{+6.55} & +1.66 & +0.50 \\
\rowcolor{gray!12}
Infinity-RoPE + ReCR
& 26.59 & +0.51 & \underline{2.01} & \underline{-0.62} & \textbf{-0.37} & +40.74 & +1.67 & +0.45 \\
\bottomrule
\end{tabular}%
}
\vspace{2pt}
\end{table*}

\section{Experiments}
\label{sec:experiments}

\subsection{Experimental Setup}
\label{subsec:exp_setup}

We evaluate text-to-video decoding. Our primary benchmark contains \(128\) prompts from the MovieGen prompt suite. For each prompt, we report both \(30\)- and \(60\)-second rollouts.
To make evaluation more diagnostic, we additionally measure drift-oriented indicators over the rollout trajectory, including text-alignment stability, aesthetic drift, color drift, sharpness drift, LPIPS-based temporal drift, and DINO-based feature drift. Details of the setup are provided in App.~\ref{app:experimental_setup}.

\subsection{Standard Long-Video Benchmarking and Qualitative Evaluation}
\label{subsec:exp_main}

We compare ReCR against both training-based and inference-time baselines. The training-oriented group includes Self-Forcing \citep{huang2025selfforcing}, Causal-Forcing \citep{zhu2026causalforcing}, Self-Forcing++ \citep{cui2025selfforcingpp}, and HiAR \citep{zou2026hiar}. The inference-time group includes Deep Forcing \citep{yi2025deepforcing} and Infinity-RoPE \citep{yesiltepe2025infinityrope}. Unless otherwise specified, all inference-time methods are reimplemented on the same \(4\)-step chunk-wise causal diffusion backbone with identical text conditioning, tokenizer, and sampling schedule. 
Table~\ref{tab:vbench_30s_60s_merged} reports the main long-video benchmark. ReCR improves the aggregated quality score across all backbone families on both \(30\)- and \(60\)-second rollouts, with mean gains of \(1.41\) and \(1.94\) points, respectively, so the average benefit grows where long-horizon degradation is stronger. Dynamic Degree also rises for every backbone family, by up to \(17.92\) points on \(30\)-second and \(23.65\) points on \(60\)-second rollouts. Table~\ref{tab:drift_abs_30s} complements the benchmark averages with a head-to-tail drift analysis on \(30\)-second videos. Here, ReCR  reduces text-alignment drop and aesthetic drift, and also moves signed drift metrics closer to zero.

Table~\ref{tab:efficiency} in App.~\ref{app:computational_resources} reports system efficiency on \(60\)-second rollouts.

\subsection{Mechanistic Analysis of Visible Context Allocation}
\label{subsec:visible_context_allocation}

\begin{figure}[t]
    \centering
    \includegraphics[width=\linewidth]{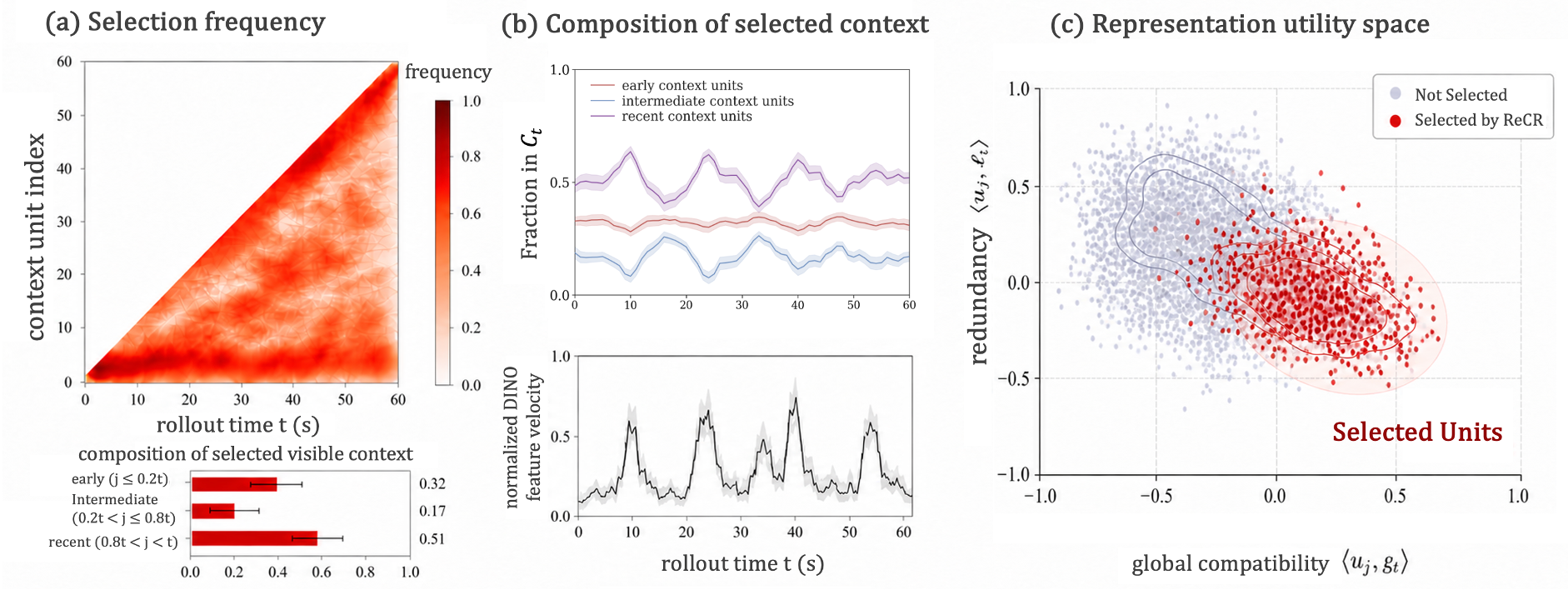}
    \caption{\textbf{Dynamic allocation of ReCR's selected visible context.}
    (a) Selection frequency over rollout time and context-unit index.
    (b) Composition of \(C_t\) and normalized DINO feature velocity over rollout.
    (c) Representation-utility space induced by \(g_t\) and \(\ell_t\), with selected units highlighted.}
    \label{fig:dynamic_vis}
\end{figure}

We examine whether ReCR constructs \(C_t\) as a genuinely selected visible context rather than a longer contiguous recent context. For each rollout step, we record the selected mask, the temporal location of selected context units in \(H_t\), and the utility components derived from \(u_j\), \(g_t\), and \(\ell_t\). Figure~\ref{fig:dynamic_vis} shows three consistent behaviors: ReCR preserves support from early and recent context units, dynamically assigns part of the budget to intermediate context units, and selects units with high compatibility to \(g_t\) while avoiding excessive redundancy with \(\ell_t\). Detailed aggregation rules are in App.~\ref{app:visible_context_diagnostics}.

\begin{figure}[t]
    \centering
    \includegraphics[width=\linewidth]{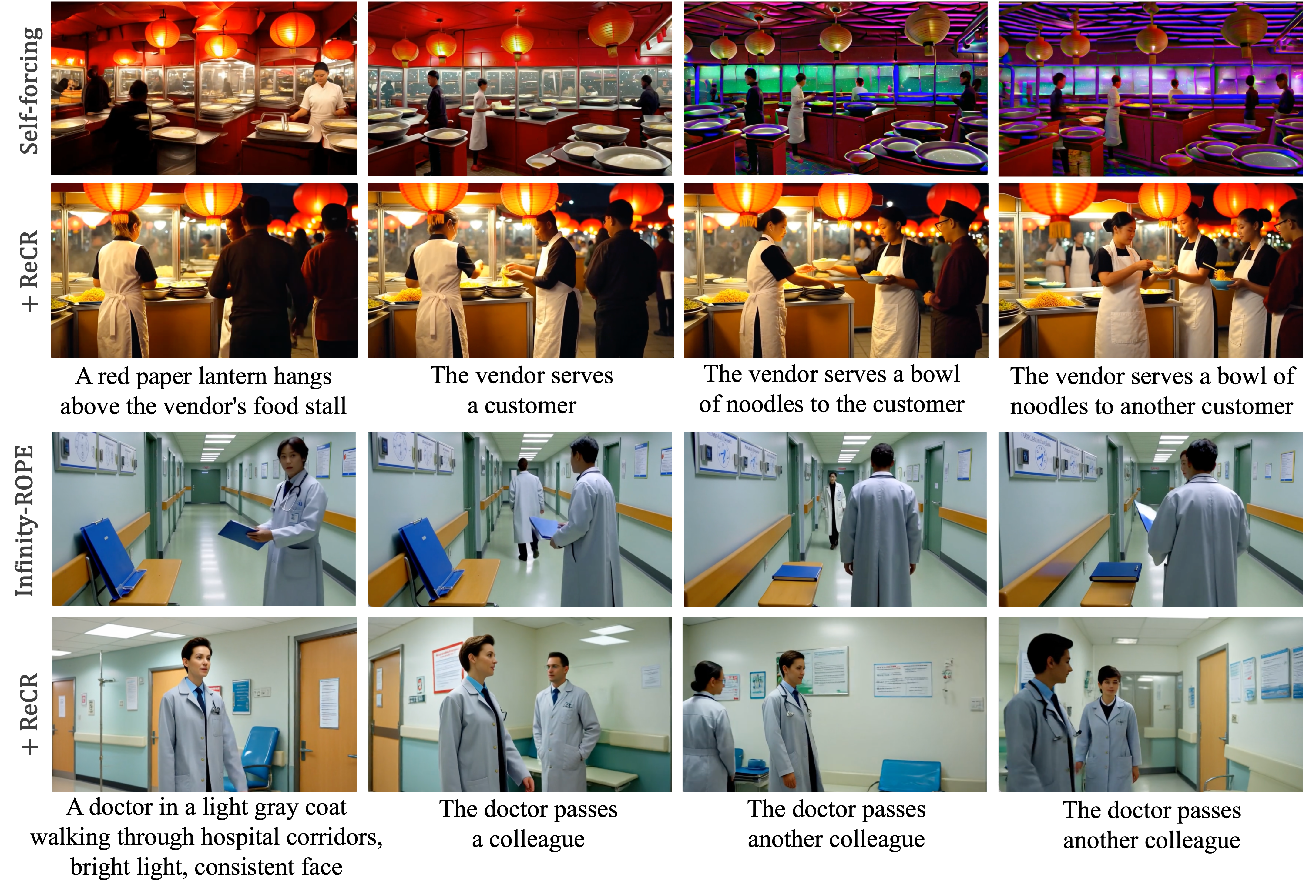}
    \caption{\textbf{Qualitative transition examples.}
    ReCR improves cross-segment consistency of persistent visual anchors and requested local actions under the same directive-switch schedules.}
    \label{fig:transition_results}
\end{figure}

\begin{figure}[t]
    \centering
    \includegraphics[width=\linewidth]{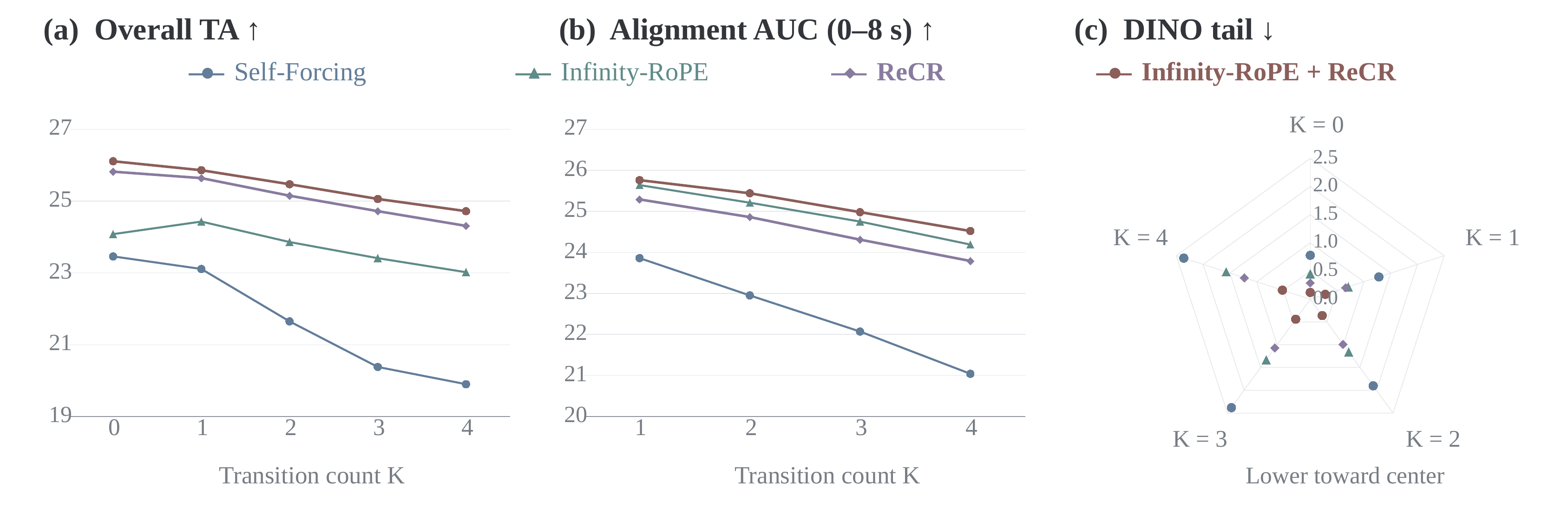}
    \caption{\textbf{Transition-count prompt-switch scaling.}
    (a) Overall text alignment (TA) to the active directive versus transition count \(K\).
    (b) Early post-switch alignment, summarized by AUC over \(0\)--\(8\) seconds, versus \(K\).
    (c) Absolute DINO tail shown as a radial dot plot: each spoke corresponds to a transition count, and distance from the center represents the metric value; closer to the center is better.
    All panels share the same method colors and markers. Higher is better in (a,b), and lower is better in (c).}
    \label{fig:transition_prompt_switch}
\end{figure}

\subsection{Transition-count Prompt Switching}
\label{subsec:transition_prompt_switch}

We test ReCR under directive changes, where the decoder must follow new instructions while preserving useful history. We build a transition-count benchmark with \(24\) prompt families. Each \(60\)-second rollout is split into five \(12\)-second directive segments, and the number of switches varies as \(K\in\{0,1,2,3,4\}\) under matched seeds, sampling schedules, and visible-token budgets. Figure~\ref{fig:transition_results} shows that ReCR variants better preserve recurring visual anchors and local motion across switches. Figure~\ref{fig:transition_prompt_switch} suggests that cache flushing and selected visible context rewriting improve complementary aspects of prompt switching. Details are given in App.~\ref{app:transition_prompt_benchmark}.

\section{Conclusion and Discussion}
\label{sec:conclusion}

We presented ReCR, a training-free decoding strategy that treats autoregressive video generation as selected visible context construction over the generated history. Starting from the question of what should remain visible under a fixed context budget, ReCR replaces the contiguous recent context with a selected visible context, scores context units, and stabilizes selection with boundary coverage, a selected-set redundancy penalty, a spatial-range exchange, and temporal reindexing. Across benchmark results, drift diagnostics, visible-context analyses, and transition-count prompt switching, ReCR improves long-horizon stability and motion evolution by changing which history the decoder reads.

ReCR suggests that autoregressive long-video decoding is not only a sampling problem or a cache-efficiency problem, but also a context-allocation problem. Existing long-context, KV-cache, sparse-attention, and memory methods make longer rollouts more feasible; ReCR adds a complementary direction by deciding which parts of the generated history should be exposed at each step. Its current limitations are also clear. ReCR depends on whether internal decoder representations can reliably separate persistent visual anchors, evolving events, and redundant local continuation; it also makes myopic stepwise decisions and operates on whole latent frames rather than object- or scene-level memory. Future work could learn adaptive utility scoring, jointly decide which history to expose and how to place it on the causal timeline, and combine ReCR-style evidence allocation with cache-based prompt responsiveness.

\newpage

\bibliographystyle{plain}
\bibliography{neurips_2025}
%%%%%%%%%%%%%%%%%%%%%%%%%%%%%%%%%%%%%%%%%%%%%%%%%%%%%%%%%%%%
\newpage

\appendix

\section{Additional Details for the ReCR Algorithm}
\label{app:recr_algorithm}

\subsection{Explicit Inference Procedure}
\label{app:recr_inference}

At the first generator pass of the block that starts at \(y_t\), ReCR constructs the selected visible context \(C_t\) from the generated history \(H_t\) as follows.

\begin{enumerate}
\item Load the stored unit representation \(u_j\) and spatial-layout statistic \(\rho_j\) of every committed context unit \(y_j \in H_t\).
\item Compute the early-biased weight \(\omega_{t,j}^{e}\) and recent-biased weight \(\omega_{t,j}^{r}\) for every \(j<t\).
\item Compute the global summary \(g_t\) and recent summary \(\ell_t\).
\item Compute the representation utility \(s_t(j)\) for each candidate context unit.
\item Select a budgeted mask \(m_t^{\star}\) by greedily maximizing the objective with representation utility, boundary coverage \(\Phi_t\), and the selected-set redundancy penalty \(\Psi_t\). When \(t-1\le B\), keep the whole history instead.
\item Form the initial selection \(\widetilde{C}_t = \{y_j \in H_t \mid m_{t,j}^{\star}=1\}\).
\item Apply the spatial-range exchange to obtain the selected visible context \(C_t\).
\item Sort the selected context units chronologically, apply the temporal reindexing map \(\pi_t\), and place the current block at positions \(N_t+1,\ldots,N_t+L\).
\item Run the denoising passes of the block with \(C_t\) and \(\pi_t\) fixed, applying RoPE to the stored keys of \(C_t\) at their compact positions, to obtain \(\hat{y}_{t:t+L-1} = D_{\theta}(\xi_t,c,C_t,\pi_t)\).
\item Run the clean context pass with the same \(C_t\) and \(\pi_t\), store the keys and values of the new units for every decoder layer, compute their \(u_j\) and \(\rho_j\) from the layer-\(\ell\) output, append them to the generated history, and continue to the next block.
\end{enumerate}

\subsection{Greedy Selection Approximation}
\label{app:recr_greedy}

The exact budgeted objective over all subsets can be expensive when \(H_t\) is long. In practice, we use a greedy approximation that follows the same scoring principle. Starting from an empty set, the algorithm repeatedly adds the candidate unit with the largest marginal gain
\[
\Delta_t(j \mid S)
=
s_t(j)
+
\mu
\left[
\Phi_t(S \cup \{j\})-\Phi_t(S)
\right]
-
\gamma
\sum_{i \in S}
\langle u_i,u_j\rangle,
\]
until \(B\) context units are selected, with ties going to the earlier unit. Here \(S\) denotes the current selected set, and \(\Phi_t(S)\) is evaluated by treating the units in \(S\) as active mask entries. This approximation preserves the main structure of ReCR: high representation utility, boundary coverage, and suppression of repeated evidence inside the selected visible context.

\subsection{Spatial-Range Exchange}
\label{app:spatial_exchange}

Let \(\bar z_j\) be the mean of the layer-\(\ell\) output tokens of \(y_j\) before normalization, taken from the same clean context pass as \(u_j\). Let \(\bar z_j^{x}\) and \(\bar z_j^{y}\) be the means of the same tokens weighted by their horizontal and vertical cell-center coordinates in \([-1,1]\). With \(\hat z_j=\bar z_j/\lVert\bar z_j\rVert_2\) and \(r_j^{a}=\bar z_j^{a}-\langle \bar z_j^{a},\hat z_j\rangle \hat z_j\) for \(a\in\{x,y\}\), the spatial-layout statistic is
\[
\rho_j
=
\frac{\sqrt{\lVert r_j^{x}\rVert_2^2+\lVert r_j^{y}\rVert_2^2}}{\lVert \bar z_j\rVert_2}.
\]
This statistic is a reading of the layout of decoder features, not a measure of motion or identity.

Let \(\rho^{-}\) and \(\rho^{+}\) be the minimum and maximum of \(\rho_j\) over \(\widetilde{C}_t\), and let \(R_t\) be the latest \(L\) history units. The exchange is considered only when \(t-1>B\), \(R_t\subseteq\widetilde{C}_t\), and \(\rho_j\) is defined for every history unit. The incoming unit \(y_{j^{+}}\) is chosen from the minimizer and maximizer of \(\rho_j\) over \(H_t\): a candidate must lie outside \(\widetilde{C}_t\) and outside \([\rho^{-},\rho^{+}]\), and the candidate farthest from this interval is taken. A selected unit \(y_i\) may leave only if \(y_i\notin R_t\), neither \(y_{i-1}\) nor \(y_{i+1}\) belongs to \(\widetilde{C}_t\), and the range of \(\rho\) over \((\widetilde{C}_t\setminus\{y_i\})\cup\{y_{j^{+}}\}\) contains \([\rho^{-},\rho^{+}]\). Among these units, the one with the smallest utility \(s_t(i)\) leaves. All ties go to the earlier unit. If there is no incoming unit or no eligible outgoing unit, \(C_t=\widetilde{C}_t\). The exchange uses stored statistics only and adds no decoder forward pass.

\subsection{Definition of Context Unit Regions}
\label{app:context_unit_regions}

For analysis and visualization, context units in \(H_t\) can be grouped by their normalized coordinate \(\tau_t(j)\). We use early context units for small \(\tau_t(j)\), recent context units for large \(\tau_t(j)\), and intermediate context units for the remaining range. These groups are used only for diagnostics and ablations. The selection rule itself does not require a hard partition of \(H_t\), because \(C_t\) is determined by \(s_t(j)\), \(\Phi_t\), \(\Psi_t\), and the spatial-range exchange.

\subsection{Backbone Compatibility}
\label{app:backbone_compatibility}

ReCR requires only reusable unit representations, stored keys and values of committed units, and control over the context units visible to the decoder. It does not update model parameters or introduce an additional training objective. For a given autoregressive video backbone, the implementation uses the same text condition, decoder input, sampling schedule, and visible context budget as the corresponding baseline, while replacing the fixed context rule with the selected visible context \(C_t\) and temporal reindexing map \(\pi_t\). The number of visible tokens never exceeds the native attention window.

To keep every past unit selectable, ReCR retains the keys and values of all committed units, before positional encoding and for every decoder layer, in a full-history bank on the GPU. The decoder reads this bank only through \(C_t\), so the bank raises peak memory, in proportion to rollout length, but not the visible-token budget.

All backbones share \(\ell=20\) (zero-indexed, out of \(30\) decoder blocks), \(\kappa=\lambda=\mu=\gamma=1\), \(\beta=4\), \(\varepsilon=10^{-6}\), and \(L=3\).

\section{Additional Experimental Details}
\label{app:experimental_details}

\subsection{Experimental Setup}
\label{app:experimental_setup}

We evaluate ReCR on long-horizon text-to-video decoding. The primary benchmark contains \(128\) prompts from the MovieGen prompt suite. For each prompt, we generate both \(30\)-second and \(60\)-second rollouts. Unless otherwise specified, all methods compared within the same backbone family use the same text condition, random seed, sampling schedule, causal decoding protocol, and visible-token budget. This matched-budget design is important because ReCR changes which generated context units are visible to the decoder, rather than increasing the number of visible tokens.

We compare against two groups of baselines. The first group contains autoregressive or causal long-video backbones, including Self-Forcing, Self-Forcing++, LongLive, Causal-Forcing, and HiAR. The second group contains training-free inference-time methods, including Deep Forcing and Infinity-RoPE. For each applicable backbone, the corresponding ReCR variant keeps the backbone parameters fixed and replaces only the visible-context construction rule with selected visible context construction and temporal reindexing, as detailed in App.~\ref{app:backbone_compatibility}. For inference-time comparisons, the backbone implementation uses the same chunk-wise causal diffusion setting, text conditioning, tokenizer, and sampling schedule as the corresponding baseline.

All reported main experiments are run on a single NVIDIA H200 GPU with batch size \(1\). We report block latency in Table~\ref{tab:efficiency}. The efficiency numbers are measured on \(60\)-second rollouts. ReCR introduces additional overhead from representation scoring, budgeted subset selection, and re-encoding the selected keys, and its full-history key--value bank accounts for its additional GPU memory, but it does not require training, fine-tuning, or additional model parameters.

\subsection{Evaluation Details}
\label{app:evaluation_details}

First, the main long-video benchmark follows the VBench-Long quality block and reports Total Score, Dynamic Degree, Subject Consistency, Background Consistency, Temporal Flickering, Motion Smoothness, Aesthetic Quality, and Imaging Quality. Table~\ref{tab:vbench_30s_60s_merged} reports each entry as \(30\mathrm{s}/60\mathrm{s}\). Higher values are better for all quality-block metrics.

Second, we compute drift-oriented diagnostics to measure how video quality changes along the rollout trajectory. We divide each generated video into temporal segments and compute head-to-tail statistics. The drift table reports semantic stability through \(\mathrm{TA}\) overall and \(\mathrm{TA}\) tail, appearance and low-level drift through aesthetic drift, aesthetic-quality signed drift, saturation shift, and sharpness tail, and temporal drift through LPIPS tail and DINO tail. Lower is better for \(\mathrm{TA}\) tail and aesthetic drift. For signed metrics and feature-tail metrics, values closer to zero indicate better stability.

For all tables, best and second-best values are highlighted following the rule stated in the table caption. For signed drift metrics, ranking is based on absolute distance to zero rather than raw numerical value. All post-processing and metric computation scripts use the same generated videos for baselines and ReCR variants, so differences come from the decoding method rather than from metric-side sampling.

\subsection{Visible Context Diagnostics}
\label{app:visible_context_diagnostics}

For the dynamic allocation analysis in Fig.~\ref{fig:dynamic_vis}, we aggregate ReCR selections over all prompts, seeds, and rollout steps. At each step \(t\), the selected visible context \(C_t\) is represented by the binary selected mask \(m_t\). We record the normalized temporal coordinate of every selected context unit within the generated history \(H_t\), together with the representation-utility components computed from the unit representation \(u_j\), the global summary \(g_t\), and the recent summary \(\ell_t\).

To analyze the temporal composition of \(C_t\), we divide context units into three diagnostic groups according to their normalized position in \(H_t\): early context units, intermediate context units, and recent context units. These groups are used only for visualization and aggregation. The ReCR selection objective itself does not impose this hard partition, since \(C_t\) is determined by representation utility \(s_t(j)\), boundary coverage \(\Phi_t\), the selected-set redundancy penalty \(\Psi_t\), and the spatial-range exchange.

Panel~(a) reports the selection frequency of each context-unit position over rollout time. The heatmap shows whether the selected visible context is concentrated near the recent boundary or distributed over the generated history. Panel~(b) reports the fraction of \(C_t\) assigned to the three diagnostic groups at each rollout step, together with normalized DINO feature velocity. We use this velocity as a proxy for motion and event-change demand. Panel~(c) projects candidate context units into the space defined by global compatibility \(\langle u_j,g_t\rangle\) and redundancy with the recent summary \(\langle u_j,\ell_t\rangle\). This view tests whether selected units follow the intended representation-utility geometry while still satisfying boundary coverage, budget, and selected-set redundancy constraints.

All curves are averaged over prompts and seeds. Shaded regions denote variability across samples. DINO feature velocity is normalized within each rollout before aggregation so that the plot reflects relative motion demand rather than differences in absolute feature scale across videos.

\subsection{Transition-Count Prompt-Switch Benchmark}
\label{app:transition_prompt_benchmark}

We build the transition-count benchmark from \(24\) controlled prompt families. Each family fixes a subject, scene, visual style, and persistent identity anchor, and provides five candidate segment directives. A \(60\)-second rollout is divided into five \(12\)-second segments. For a directive sequence \(q=(q_1,\ldots,q_5)\), the transition count is
\[
K(q)=\#\{r\in\{1,2,3,4\}:q_r\neq q_{r+1}\}.
\]
For each \(K\), we choose \(K\) switched boundaries among the four possible segment boundaries. With \(24\) prompt families, the switch locations are balanced: for \(K=1\) and \(K=3\), each of the four boundary patterns appears six times; for \(K=2\), each of the six boundary pairs appears four times; \(K=0\) and \(K=4\) are deterministic. Given a boundary set \(\mathcal{B}_K\), we start from the first directive and advance to the next unused directive only when a switched boundary is crossed. Formally, \(z_1=1\), \(z_{r+1}=z_r+\mathbf{1}[b_r\in\mathcal{B}_K]\), and \(q_r=D_{z_r}\). The actual text condition for segment \(S_r\) is the shared family anchor concatenated with the active directive \(q_r\). This keeps subject, scene, and style fixed while varying only the number and placement of directive switches.

For Fig.~\ref{fig:transition_prompt_switch}a, \(\mathrm{TA}\) overall is computed by applying the same text-image alignment pipeline to each segment using its active directive and averaging over the five segments. For Fig.~\ref{fig:transition_prompt_switch}b, we align frames after each switch to the new active directive and plot the early post-switch AUC over \(0\)--\(8\) seconds against transition count \(K\in\{1,2,3,4\}\). No post-switch AUC is shown for \(K=0\), since no directive switch occurs. For Fig.~\ref{fig:transition_prompt_switch}c, \(\mathrm{DINO~tail}\) is computed with the same tail-drift protocol used in Table~\ref{tab:drift_abs_30s}, and we plot its absolute value because values closer to zero indicate less feature drift. All methods are evaluated with the same prompt schedules, random seeds, sampling schedule, and visible-token budget.
\subsection{Computational Resources}
\label{app:computational_resources}

All experiments reported in the paper were conducted on a single NVIDIA H200 GPU. We use batch size \(1\) for long-video generation because autoregressive \(60\)-second rollouts require maintaining the generated history, decoder-visible context, and intermediate activations over extended horizons. The block-latency measurements in Table~\ref{tab:efficiency} are obtained from the same hardware setting.

\begin{table}[htbp]
\centering
\caption{System efficiency on \(60\)-second rollouts.}
\label{tab:efficiency}
\setlength{\tabcolsep}{12pt}
\renewcommand{\arraystretch}{1.08}
\small
\begin{tabular}{lc}
\toprule
\textbf{Method} & \textbf{Block Latency} $\downarrow$ \\
\midrule
\multicolumn{2}{l}{\textbf{Native Backbones}} \\
\midrule
Self-Forcing \citep{huang2025selfforcing}
& 728 ms \\

LongLive \citep{yang2025longlive}
& 685 ms \\

Causal-Forcing \citep{zhu2026causalforcing}
& 737 ms \\

Infinity-RoPE \citep{yesiltepe2025infinityrope}
& 667 ms \\
\midrule
\multicolumn{2}{l}{\textbf{Backbone + ReCR}} \\
\midrule
\rowcolor{gray!12}
Self-Forcing + ReCR
& 712  ms \\

\rowcolor{gray!12}
LongLive + ReCR
& \textbf{659 ms} \\

\rowcolor{gray!12}
Causal-Forcing + ReCR
& 711 ms \\

\rowcolor{gray!12}
Infinity-RoPE + ReCR
& 710 ms \\
\bottomrule
\end{tabular}
\end{table}

\section{Additional Experimental Results}
\label{app:additional_experimental_results}

\begin{figure}
    \centering
    \includegraphics[width=\linewidth]{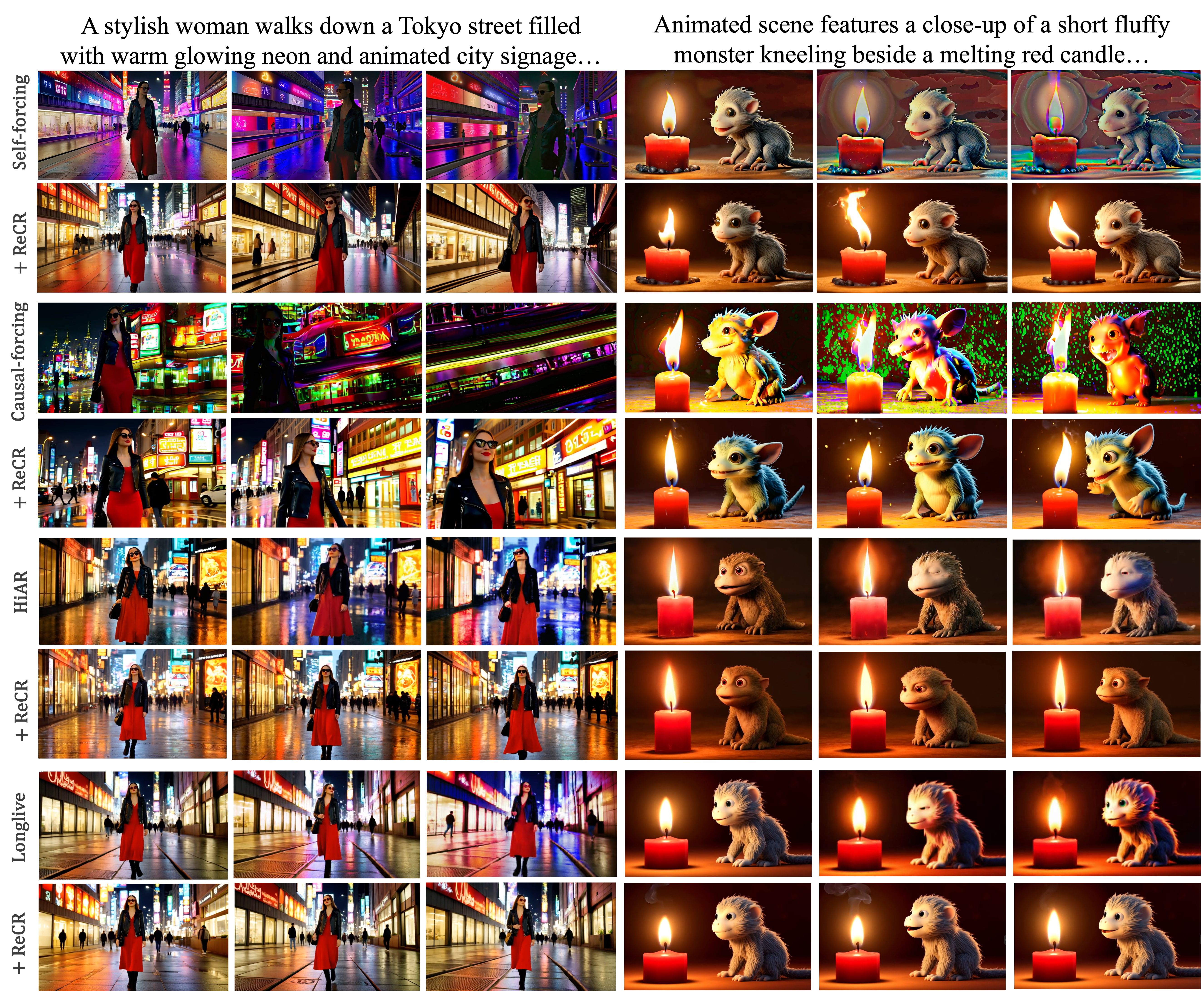}
    \caption{ReCR's results on more backbone and more frame.}
    \label{fig:ReCR_teaser_training_result}
\end{figure}

\paragraph{Mechanistic Robustness Under Perturbation}
\label{app:mechanistic_robustness}

\begin{figure}
    \centering
    \includegraphics[width=\linewidth]{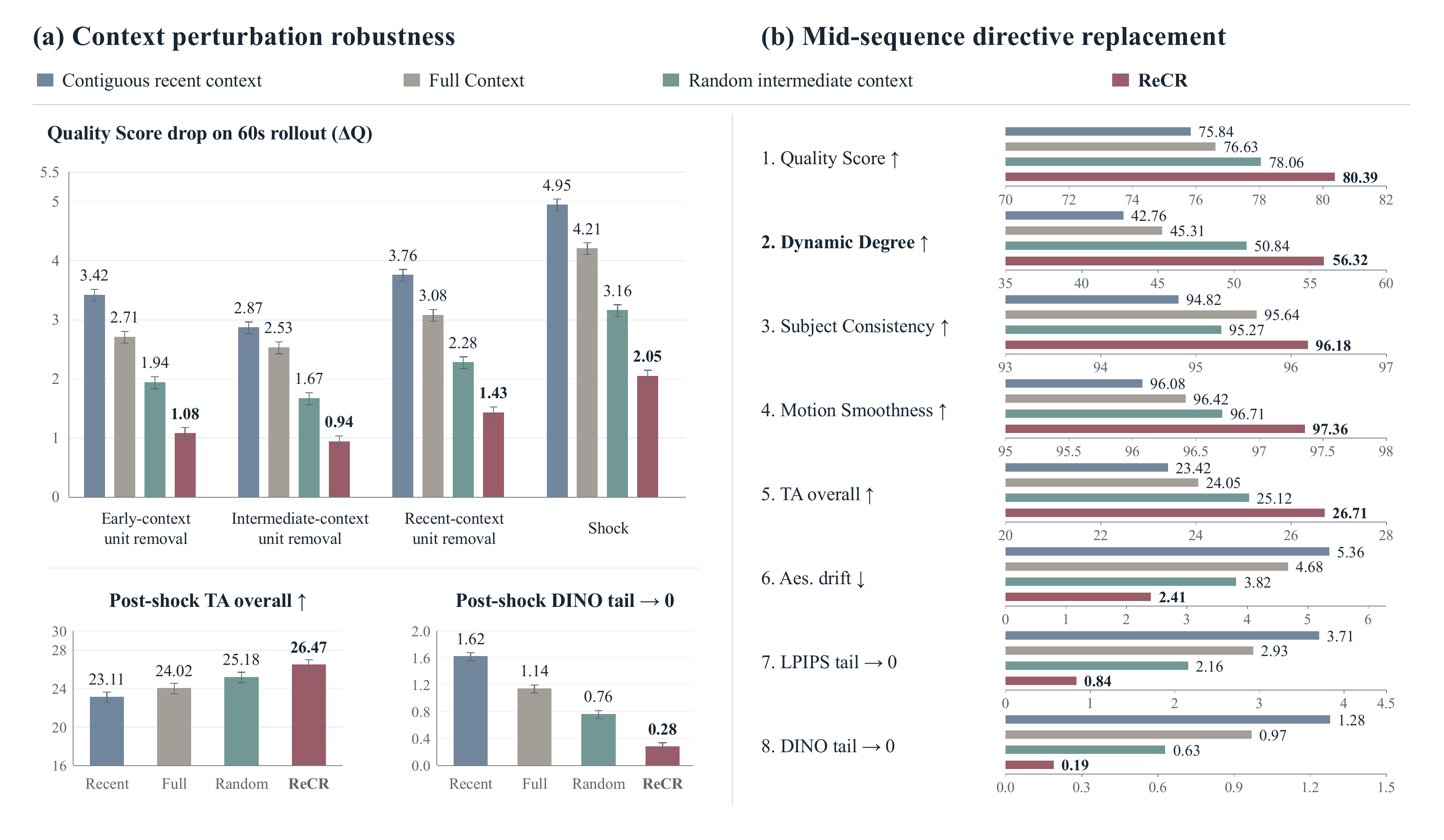}
    \caption{\textbf{Mechanistic robustness analysis.}
    We compare four context construction rules under controlled perturbations: contiguous recent context, Full Context, random intermediate context, and ReCR.}
    \label{fig:mechanistic}
\end{figure}

This analysis tests whether the gains of ReCR come from a specific context-allocation mechanism rather than from simply exposing more history. We compare ReCR with three controlled alternatives. The contiguous recent context baseline exposes only the most recent generated units. Full Context exposes the generated history as broadly as the budget allows. The random intermediate context baseline samples non-recent context units without representation-guided scoring. ReCR uses the full selection objective with representation utility, boundary coverage, selected-set redundancy penalty, and temporal reindexing.

We perturb the visible history in four ways. Early-context-unit removal tests whether a method depends on units that stabilize persistent identity and scene layout. Intermediate-context-unit removal tests sensitivity to units that may carry event-evolution evidence. Recent-context-unit removal tests sensitivity to short-range temporal continuity. Shock applies a stronger one-shot perturbation and measures recovery afterward. For each method, we report the quality-score drop on \(60\)-second rollouts, post-shock text alignment, and post-shock DINO tail. Lower quality-score drop and DINO tail indicate stronger robustness, while higher post-shock text alignment indicates better recovery.

The results show that ReCR is not equivalent to retaining the full history or sampling intermediate units at random. Fixed recent-context rules are vulnerable when local continuity is perturbed, full-context rules can still expose redundant or stale evidence, and random intermediate context lacks a mechanism for selecting useful evidence. ReCR is consistently more stable because it selects context units according to their representation utility while maintaining boundary coverage and suppressing repeated evidence inside the selected set.

%\subsection{Ablation on $\mu$ and $\gamma$}

\section{Limitations}
\label{app:limitations}

ReCR improves long-horizon decoding by deciding which generated context should remain visible, but it also has some limitations. First, the method depends on the quality of internal decoder representations. If the unit representations do not separate persistent visual anchors, evolving events, and redundant local continuation, the selected visible context may not reflect the evidence that is actually useful for future prediction. This limitation is more likely to appear in highly abstract prompts, rapidly changing scenes, or videos where the relevant evidence is spatially localized but weak at the unit level.

Second, ReCR makes local stepwise decisions. At each rollout step, it selects context units for the next prediction, but it does not explicitly plan which evidence may become important several segments later. This design keeps the method training-free and lightweight, but it may be suboptimal for long stories with delayed causal dependencies, repeated scene returns, or objects that disappear and reappear after long gaps.

\section{Future Work}
\label{app:future_work}

The main direction suggested by ReCR is to make the generated history a structured resource for future decoding. One natural extension is to learn adaptive utility scoring from rollout feedback. The current representation utility is hand-designed to balance compatibility with the broader rollout state against redundancy with the recent continuation. A learned scorer could adapt this balance across prompt types, motion patterns, and backbone architectures while preserving the same selected-visible-context formulation.

A second direction is predictive context planning. ReCR currently selects the context for the next decoding step, but long videos often require evidence that remains useful across many future steps. Future methods could select context not only for immediate prediction, but also for anticipated future needs, such as returning to a previous scene, maintaining a recurring object, or preserving a long-range action plan.

A third direction is finer-grained memory. ReCR selects whole latent frames as context units, which is simple and backbone-compatible. More structured variants could select objects, regions, motion tracks, or geometry-aware scene elements. Such memory could help when the important evidence occupies only a small part of a frame or when multiple objects evolve independently.

Finally, ReCR could be combined more tightly with cache-based prompt responsiveness. The transition-count experiments suggest that cache flushing and selected visible context rewriting address different failure modes: cache mechanisms help the decoder respond quickly to new directives, while ReCR improves which generated evidence remains available after the switch. A hybrid system could jointly decide when to flush, what to retain, and how to expose the retained history on the causal timeline.

\section{Existing Assets and Licenses}
\label{app:asset_licenses}

This work uses existing research assets only for evaluation, comparison, and implementation of baselines. We use the MovieGen prompt suite for text-to-video prompts and cite the original MovieGen paper in the main text. We evaluate generated videos with standard long-video and drift metrics, including VBench-Long-style quality metrics and commonly used feature backbones or perceptual measures such as DINO, LPIPS, and text-image alignment models. We also compare against published autoregressive video generation and training-free inference methods, including Self-Forcing, Self-Forcing++, LongLive, Causal-Forcing, HiAR, Deep Forcing, and Infinity-RoPE, with citations provided in the main paper.

We do not claim ownership of these existing assets. All original datasets, prompts, codebases, model checkpoints, and evaluation tools remain subject to their original licenses and terms of use. Our use is limited to research evaluation and method comparison. When releasing code, we will provide clear instructions for users to obtain any third-party assets from their original sources and to follow the corresponding license terms. The released package will not redistribute third-party model weights or datasets unless their licenses explicitly permit redistribution.

The new asset introduced by this work is the ReCR implementation and associated evaluation scripts. The code is released through the repository listed in the abstract. The release includes the context selection implementation, metric reductions, tests, and instructions for integrating ReCR with a video backbone; backbone-specific integrations are not included. Generated videos shown in the paper are produced for research demonstration and are not intended as a standalone dataset for downstream training.

\section{Broader Impact}
\label{app:broader_impact}

ReCR is a decoding method for improving long-horizon autoregressive video generation. Its positive impact is to make long-video generation more stable under fixed visible-token budgets, which can support research on controllable video synthesis, simulation, creative tools, and diagnostic evaluation of autoregressive decoders. By focusing on test-time context allocation rather than retraining, ReCR may also make long-video improvements more accessible to researchers who cannot afford large-scale model training.

The same improvement in long-video coherence can also increase the quality of synthetic media. More stable long videos may be misused for misleading content, impersonation, or disinformation when combined with unsafe prompts, identity-conditioned generation, or downstream editing tools. ReCR itself does not introduce a new dataset, identity model, or deployment system, but it can improve the outputs of existing video generators. Responsible use therefore requires the same safeguards expected for generative video systems, including provenance tracking, watermarking when available, prompt and content filtering, and clear disclosure when generated videos are used in public-facing settings.

Our experiments use controlled text prompts and focus on generic scenes rather than real private individuals. The method is intended for research use and analysis of autoregressive decoding behavior. We do not recommend deploying ReCR-enhanced generation systems without the safety mechanisms, content policies, and misuse monitoring appropriate for the underlying video generator.

\section{Use of Large Language Models}
\label{app:llm_usage}

We used large language models only for editing and improving the clarity of the manuscript, including grammar, spelling and word choice,.

Specifically, we used Gemini 2.5 Pro and Gemini 3 Pro to polish early manuscript drafts. Example prompt include:

\begin{quote}
\textit{Please check the grammar, spelling, word choice, and sentence flow of the following LaTeX text. Keep all equations, labels, references, and macros unchanged, and avoid changing the scientific content.}
\end{quote}

\newpage
\section*{NeurIPS Paper Checklist}

\begin{enumerate}

\item {\bf Claims}
    \item[] Question: Do the main claims made in the abstract and introduction accurately reflect the paper's contributions and scope?
    \item[] Answer: \answerYes{}
    \item[] Justification: The abstract and introduction state that ReCR is a training-free method for representation-guided context rewriting under a fixed visible-context budget. These claims are supported by the method section, standard long-video benchmarks, drift diagnostics, visible-context analysis, and transition-count prompt-switch experiments.

\item {\bf Limitations}
    \item[] Question: Does the paper discuss the limitations of the work performed by the authors?
    \item[] Answer: \answerYes{}
    \item[] Justification:The paper discusses limitations in the Conclusion and Discussion and in App.~\ref{app:limitations}. These include dependence on internal decoder representations, local stepwise selection, coarse unit-level context granularity, and additional inference-time overhead.

\item {\bf Theory assumptions and proofs}
    \item[] Question: For each theoretical result, does the paper provide the full set of assumptions and a complete (and correct) proof?
    \item[] Answer: \answerNA{}
    \item[] Justification: The paper does not present formal theoretical theorems or proofs. The mathematical content consists of definitions, scoring functions, and the ReCR selection objective used for the decoding algorithm.

    \item {\bf Experimental result reproducibility}
    \item[] Question: Does the paper fully disclose all the information needed to reproduce the main experimental results of the paper to the extent that it affects the main claims and/or conclusions of the paper (regardless of whether the code and data are provided or not)?
    \item[] Answer:  \answerYes{}
    \item[] Justification: The paper provides the ReCR objective, explicit inference procedure, greedy selection approximation, experimental setup, evaluation metrics, visible-context diagnostics, and transition-count benchmark construction. The core algorithm is provided through the repository linked in the abstract.

\item {\bf Open access to data and code}
    \item[] Question: Does the paper provide open access to the data and code, with sufficient instructions to faithfully reproduce the main experimental results, as described in supplemental material?
    \item[] Answer:  \answerYes{}
    \item[] Justification: The repository linked in the abstract provides the core ReCR algorithm, evaluation reductions, tests, and integration instructions. Backbone-specific integrations are not included.

\item {\bf Experimental setting/details}
    \item[] Question: Does the paper specify all the training and test details (e.g., data splits, hyperparameters, how they were chosen, type of optimizer) necessary to understand the results?
    \item[] Answer:  \answerYes{}
    \item[] Justification: The paper specifies the MovieGen-128 prompt benchmark, 30- and 60-second rollout settings, compared backbone families, matched visible-token budget, shared seeds and sampling schedules, drift metrics, and transition-count benchmark. Additional details are provided in App.~\ref{app:experimental_details}.

\item {\bf Experiment statistical significance}
    \item[] Question: Does the paper report error bars suitably and correctly defined or other appropriate information about the statistical significance of the experiments?
    \item[] Answer: \answerNo{}
    \item[] Justification: The main quantitative tables report aggregate benchmark scores and drift metrics, and the diagnostic plots include variability visualization where applicable. However, due to computational constraint, the current version does not provide confidence intervals or formal statistical significance tests for main benchmark tables.

\item {\bf Experiments compute resources}
    \item[] Question: For each experiment, does the paper provide sufficient information on the computer resources (type of compute workers, memory, time of execution) needed to reproduce the experiments?
    \item[] Answer: \answerYes{}
    \item[] Justification: The paper reports that all experiments are run on a single NVIDIA H200 GPU and provides block latency in Table~\ref{tab:efficiency}. App.~\ref{app:computational_resources} further describes the single-GPU inference setting and the additional overhead introduced by representation scoring and subset selection.
    
\item {\bf Code of ethics}
    \item[] Question: Does the research conducted in the paper conform, in every respect, with the NeurIPS Code of Ethics \url{https://neurips.cc/public/EthicsGuidelines}?
    \item[] Answer:  \answerYes{}
    \item[] Justification: The research conforms to the NeurIPS Code of Ethics. The work uses controlled text prompts, evaluates generated videos, and includes discussion of societal risks, safeguards, and existing assets.

\item {\bf Broader impacts}
    \item[] Question: Does the paper discuss both potential positive societal impacts and negative societal impacts of the work performed?
    \item[] Answer:  \answerYes{}
    \item[] Justification: App.~\ref{app:broader_impact} discusses positive impacts such as improved long-video stability for research and creative tools, as well as negative impacts such as higher-quality synthetic media misuse, impersonation, and disinformation. It also discusses mitigation strategies including provenance tracking, watermarking, content filtering, and disclosure.
    
\item {\bf Safeguards}
    \item[] Question: Does the paper describe safeguards that have been put in place for responsible release of data or models that have a high risk for misuse (e.g., pre-trained language models, image generators, or scraped datasets)?
    \item[] Answer: \answerNA{}.
    \item[] Justification: Although ReCR does not release a new generative model or dataset, the paper still discusses safeguards relevant to improved video generation.

\item {\bf Licenses for existing assets}
    \item[] Question: Are the creators or original owners of assets (e.g., code, data, models), used in the paper, properly credited and are the license and terms of use explicitly mentioned and properly respected?
    \item[] Answer: \answerYes{}
    \item[] Justification: The paper identifies the existing assets used for prompts, evaluation, baselines, and feature metrics.

\item {\bf New assets}
    \item[] Question: Are new assets introduced in the paper well documented and is the documentation provided alongside the assets?
    \item[] Answer:  \answerYes{}
    \item[] Justification: The new asset introduced by this work is the ReCR implementation and associated evaluation scripts, released through the repository linked in the abstract. 

\item {\bf Crowdsourcing and research with human subjects}
    \item[] Question: For crowdsourcing experiments and research with human subjects, does the paper include the full text of instructions given to participants and screenshots, if applicable, as well as details about compensation (if any)? 
    \item[] Answer: \answerNA{}
    \item[] Justification: The paper does not report crowdsourcing or research with human subjects.

\item {\bf Institutional review board (IRB) approvals or equivalent for research with human subjects}
    \item[] Question: Does the paper describe potential risks incurred by study participants, whether such risks were disclosed to the subjects, and whether Institutional Review Board (IRB) approvals (or an equivalent approval/review based on the requirements of your country or institution) were obtained?
    \item[] Answer: \answerNA{}
    \item[] Justification: The paper does not report research with human subjects.

\item {\bf Declaration of LLM usage}
    \item[] Question: Does the paper describe the usage of LLMs if it is an important, original, or non-standard component of the core methods in this research? Note that if the LLM is used only for writing, editing, or formatting purposes and does \emph{not} impact the core methodology, scientific rigor, or originality of the research, declaration is not required.
    %this research? 
    \item[] Answer: \answerNA{}.
    \item[] Justification: LLMs were used only for manuscript spelling and word choice. 

\end{enumerate}
%%%%%%%%%%%%%%%%%%%%%%%%%%%%%%%%%%%%%%%%%%%%%%%%%%%%%%%%%%%%

\end{document}